\documentclass[twoside]{article}

\usepackage{hyperref}
\usepackage{url}
\usepackage{graphicx}
\usepackage{subcaption}
\usepackage[round]{natbib}
\usepackage[accepted]{aistats2026}
\usepackage{amssymb}
\usepackage{fontawesome5}
\DeclareMathOperator*{\argmin}{argmin}

\begin{document}



\twocolumn[
\aistatstitle{Patch2Loc: Learning to Localize Patches for Unsupervised Brain Lesion Detection}

\aistatsauthor{
Hassan Baker \And Austin J. Brockmeier
}

\aistatsaddress{
% Department of Electrical \& Computer Engineering \\
University of Delaware \\
% Newark, DE 19716, USA \\
\texttt{\{bakerh,ajbrock\}@udel.edu}
% \And
% Dept. of Electrical \& Computer Engineering \\
% Dept.\ of Computer \& Information Sciences \\
% University of Delaware \\
% Newark, DE 19716, USA \\
% \texttt{ajbrock@udel.edu}
}
]
\begin{abstract}
Detecting brain lesions as abnormalities observed in magnetic resonance imaging (MRI) is essential for diagnosis and treatment. In the search of abnormalities, such as tumors and malformations, radiologists may benefit from computer-aided diagnostics that use computer vision systems trained with machine learning to segment normal tissue from abnormal brain tissue. While supervised learning methods require annotated lesions, we propose a new unsupervised approach (Patch2Loc) that learns from normal patches taken from structural MRI. We train a neural network model to map a patch back to its  spatial location within a slice of the brain volume. During inference, abnormal patches are detected by the anomaly score based on the error and variance of the location prediction. By applying the network in a convolutional manner, this generates a pixel-wise heatmap of anomalies providing finer-grained segmentation. We demonstrate the ability of our model to segment abnormal brain tissues by applying our approach to the detection of tumor tissues in MRI on T2-weighted images from BraTS2021 and MSLUB datasets and T1-weighted images from ATLAS and WMH datasets. We show that it outperforms the state-of-the art in unsupervised segmentation.
\end{abstract}

\faGithub\ \url{https://github.com/bakerhassan/Patch2Loc}

\section{INTRODUCTION}

Detecting and localizing abnormal brain tissue in neuroimages is a critical diagnostic task, where early intervention can mitigate severe outcomes like those from incipient tumors or cortical malformations associated with epilepsy \citep{josephson2012seizure}. While magnetic resonance imaging (MRI) is a powerful non-invasive modality for this task, modern supervised deep learning models are hindered by the scarcity of annotated data, particularly for rare and structurally diverse conditions \citep{busby2018bias, hagens2019impact}. This data bottleneck, combined with a shortage of neuroradiology experts \citep{merewitz2006portrait}, makes the development of robust unsupervised methods for anomaly detection an essential clinical and research goal.

Current state-of-the-art unsupervised approaches, including autoencoders~\citep{zimmerer2018context,sato2019predictable,Meissen2022Aug,Meissen2022Jul,Behrendt2022Apr,Bercea2025Feb}, adversarial autoencoders~\citep{brock2018large,chen2018unsupervised,baur2019deep}, denoising autoencoders~\citep{kascenas2022denoising}, transformers~\citep{pinaya2022unsupervised} and denoising diffusion probabilistic models (DDPMs)~\citep{wyatt2022anoddpm, pinaya2022fast,liang2023modality,Bercea2025Feb,behrendt2024patched,behrendt2025guided}, operate on a principle of \textbf{global} reconstruction. These methods learn the typical structure of a normal brain and flag anomalies as regions where the model fails to reconstruct the input accurately. However, this reliance on global context might provide enough clues to reconstruct the abnormal regions~\citep{wyatt2022anoddpm,Bercea2023Apr,Kascenas2023Dec}. The best-performing models are DDPMs, but they suffer from a difficult trade-off called the \textbf{`noise paradox'}~\citep{Kascenas2023Dec, bercea2023mask}—where enhancing anomaly signals can degrade the reconstruction of normal tissue, leading to false positives and limiting clinical reliability. These methods are thus constrained by a delicate balancing act of an inference-time hyperparameter (i.e., the amount of added noise) to be effective. %A more comprehensive discussion of the related work can be found in appendix~\ref{sec:related_work}.

To overcome these limitations, we propose Patch2Loc, a novel framework that fundamentally shifts the paradigm from global reconstruction to local structural assessment. Instead of learning to reconstruct an entire image, Patch2Loc is trained on a simple yet powerful localizing task: predicting the spatial coordinates of an isolated image patch from a normal brain. The intuition is that the model learns the distinct anatomical patterns characteristic of each location. When presented with a patch containing an anomaly, its structure deviates from the norm, causing the model to predict its location with high error and high uncertainty. This provides a direct and robust signal for anomaly detection based on localized structural failure. This is a key distinction from other location-based self-supervised learning (SSL) tasks such as Jigsaw  \citep{noroozi2016unsupervised}, patch ordering and context prediction  \citep{doersch2015unsupervised,NEURIPS2020_d2dc6368}, where the objective is to learn from the relationships between multiple patches (e.g., context or relative position). These tasks use the relationship between two or more patches within the global context. Their task losses are not localized to a single patch; thus, they were not designed for localized abnormality detection. In contrast, Patch2Loc can be applied convolutionally using overlapping patches to localize abnormalities by increases in the location prediction error and the model's uncertainty.

In summary, Patch2Loc offers several key advantages:

\begin{enumerate}
\item \textbf{Spatially-Aware Local Feature Learning:} By learning to predict a patch's spatial origin, Patch2Loc encodes local anatomical patterns. This allows it to effectively distinguish abnormal patches based on deviations in location prediction, directly targeting the source of the anomaly.

\item \textbf{Uncertainty-Guided Anomaly Detection:} The variance in the model's predictions serves as a powerful uncertainty estimate. Combining the location prediction error with this uncertainty creates a abnormality score that better correlates with ground truth annotations.

\item \textbf{Minimal Inference Hyperparameter Dependence:} Unlike reconstruction-based methods that require careful tuning of noise levels during inference, Patch2Loc operates with a stable configuration, providing a pixel-level abnormality heatmap without complex post-training adjustments.
\end{enumerate}

By focusing on local features and leveraging both prediction error and uncertainty, Patch2Loc offers an intuitive, interpretable, and effective solution for unsupervised anomaly detection with high potential for clinical applicability.

\begin{figure*}[t]
\centering
\includegraphics[width=0.94\textwidth]{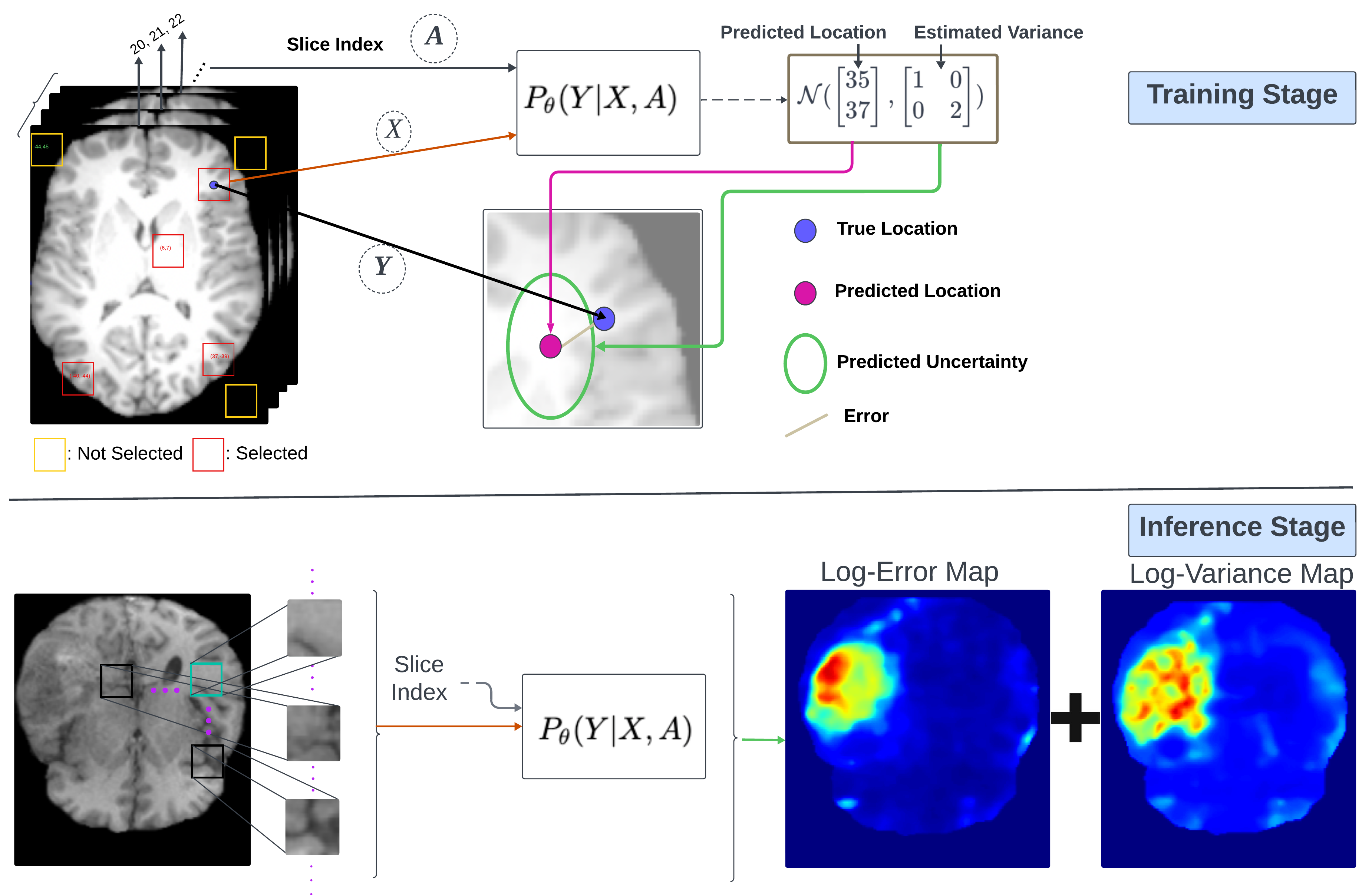}
\caption{Schematic diagram for Patch2Loc. (Top) Training stage:  A randomly selected patch from a normal MRI slice is paired with its two-dimensional location $Y$ and slice index $A$.  The relationship between the image patch, slice, and its location is modeled as a conditional distribution $P_\theta(Y|X,A)$ using a 2D Gaussian distribution defined by the mean and the variance of each coordinate that are functions of the patch and slice index. Note that patches with less than $20\%$ of its content as brain tissues, as in the yellow patch, are rejected. (Bottom)  Inference stage: Overlapping patches with a fixed stride are extracted as in convolution from an MRI slice and the model is applied to each patch. The squared norm of the error between the model's predicted mean and the patch's true location creates an error map. Likewise, the sum of the variances create a variance map. Together the sum of the logarithms of the errors and variances highlight anomalous areas.}
\label{fig:schematic}
\end{figure*}

\section{METHODOLOGY}
\label{sec:methodology}
The idea underlying Patch2Loc is that there is a strong relationship between patch content and location, due to the normal anatomical patterns in structural neuroimages, that is used by neuroradiologists when identifying structural abnormalities. With sufficient sampling across the population, this relationship can be modeled using machine learning. As each brain has slightly different sizes, the images are first registered using rigid transformation, such that the patches taken from the same coordinates contain similar anatomy.  Figure~\ref{fig:schematic} shows a summary schematic diagram for methodology behind Patch2Loc. 

\subsection{Problem Formulation}
Let  $(Y_1,Y_2,A)\in [0,100]\times [0,100] \times [0,100]$ denote the 3D location of a rectangular image patch $X\in\mathcal{X} \subset \mathbb{R}_{\ge 0}^{S_1 \times S_2}$ within the brain taken at the 2D location $Y=(Y_1,Y_2)\in [0,100]\times [0,100]$  from the slice located at $ A\in [0,100]$. The coordinates are percentages of the patch's location in absolute coordinates $(L_1,L_2,L_3)$ relative to the brain volume's spatial extent along each axis $(E_1, E_2, E_3)$:  $Y_1=100
\cdot \frac{L_1}{E_1}$, ${Y}_2=100
\cdot \frac{L_2}{E_2}$, and ${A}=100
\cdot\frac{L_3}{E_3}$. With registered brain scans, the spatial extents $E_1,E_2,E_3$ are constants common to all scans. Each 2D patch is rectangular with an absolute size of $S_1\times S_2$ chosen based on a fixed relative proportion $r=\frac{S_1}{E_1}=\frac{S_2}{E_2}$. The choice of $r$ controls the patch's coverage.

The patch and location can be described as continuous random variables jointly distributed $(X,Y,A)\sim P$. Intuitively, $P(X|Y_1=y_1,Y_2=y_2,A=a)$ is the distribution of images at a particular location $(y_1,y_2,a)$, but this is a high-dimensional distribution that is difficult to model. To capture the shared information between the image patch and its location, we model the conditional distribution $P(Y|X,A)$, which is two dimensional. For simplicity, we model this as a 2D Gaussian distribution,
\begin{equation}
P_\theta(Y|X,A) =\mathcal{N}(\mu^\theta(X,A),\Sigma^\theta(X,A)),
\end{equation}
where $\mu^\theta(X,A)$ is the 2D mean and $\Sigma^\theta(X,A)$ is the covariance matrix, both are functions of the patch $X$ and the slice location $A$, parameterized by $\theta$. To further simplify the model, we consider a diagonal covariance matrix described by the variance of the two coordinates. The model is then $Y_i | X,A \sim \mathcal{N}(\mu_i^\theta(X,A),\Sigma_{ii}^\theta(X,A)),\quad i\in\{1,2\}$. 

The functions predicting the conditional mean and variances $\mu^\theta$ and $\Sigma^\theta$ of location given the patch are modeled using a neural network with parameters $ \theta = (\theta_0, \theta_m, \theta_v)$, where  $\theta_0$ denotes the parameters of the shared patch encoder $\phi^{} :\mathcal{X}\rightarrow \mathbb{R}^d$, $\theta_m$ denotes the parameters of the mean prediction head $\tilde{\mu}^{}$, such that $\mu^\theta(X,A) = \tilde{\mu}^{}(\phi^{}(X),A)\in \mathbb{R}^2$, and $\theta_v$ denotes the parameters of log-variance head $\tilde{\varsigma}^{}$, such that $\varsigma^{\theta}(X,A)=\tilde{\varsigma}^{}(\phi^{}(X),A)\in \mathbb{R}^2$ and $\Sigma_{ii}^\theta(X,A)= \mathrm{exp}(\varsigma_i^{\theta}(X,A)),\quad i\in\{1,2\}$.

\subsection{Loss Function}
The model parameters could be  optimized to minimize the negative-log likelihood (NLL) 
\begin{equation}
\theta^* \in \argmin_\theta \mathbb{E}_{(X,Y,A)\sim P } [- \log P_\theta(Y|X,A)].    
\end{equation}
However, using NLL for estimating a model of both the mean and the variance is problematic as the component of the loss with respect to the mean estimates are scaled by the variance estimates, 
\begin{equation}
-\log P_\theta(Y|X,A)\propto \sum_{i=1}^2 \frac{\lvert Y_i - \mu_i^\theta(X,A)\rvert^2}{\mathrm{exp}(\varsigma_i^{\theta}(X,A))}+\varsigma_{i}^\theta(X,A),    
\end{equation}
which causes convergence to local minima with poor performance  \citep{nix1994estimating, seitzer2022on}. Hence, we adapt the $\beta$-NLL loss proposed in the work by \cite{seitzer2022on} that scales each dimension of the loss by  the variance estimate taken to the $\beta$th power, $\beta\in(0,1]$,
\begin{equation}
\label{eq:loss}
 \sum_{i=1}^2 \lfloor \mathrm{exp}( \varsigma_i^{\theta}(X,A)) \rfloor^\beta \left( \frac{\lvert Y_i - \mu_i^\theta(X,A)\rvert^2}{\mathrm{exp}(\varsigma_i^{\theta}(X,A))}+\varsigma_{i}^\theta(X,A)\right),    
\end{equation}
where $\lfloor \cdot \rfloor$ indicates the stop-gradient operation such that gradients are not taken with respect to this scaling. This scaling mitigates the effect of the variance estimate on the gradient of the loss with respect to the mean. Notably, when $\beta=1$ then $\mu_i^\theta$ is updated based on mean-squared error. Thus, we choose the suggested value of $\beta=0.5$  \citep{seitzer2022on}. %We confirmed that using NLL as a loss directly results in a local minimum where the model produces a high variance estimate and the mean has a high error.

\subsection{Abnormality Score}
\label{sec:abnromality_score}
To detect abnormalities with Patch2Loc, we calculate an abnormality score from the location prediction error and the sum of the variance estimates, after a log transformation.  The log-error is computed as $\log(\lVert Y - \mu^\theta(X,A)\rVert_2^2+\varepsilon)$, where $\varepsilon=0.5$ such that it does not effect large errors. The log-variance is computed as $\frac{1}{2}\log(\mathrm{det}(\Sigma^\theta(X,A)))=\frac{1}{2}\sum_{i=1}^2 \log(\Sigma^\theta_{ii}(X,A)))=\frac{1}{2}\sum_{i=1}^2 \varsigma_i^\theta (X,A)$. The log-variance captures the model's uncertainty since  adding $1+\log(2\pi)$ to the log-variance is the entropy of $\mathcal{N}(\mu^\theta(X,A),\Sigma^\theta(X,A))$. The abnormality score is
\begin{equation}
    \begin{split}
        \textrm{Score}(X,Y,A) &= \underbrace{\log(\lVert Y - \mu^\theta(X,A)\rVert_2^2+\varepsilon)}_{\text{Log(Error\textsuperscript{2})}} \\
        & \quad + \underbrace{\frac{1}{2}\varsigma_1^\theta(X,A)+ \frac{1}{2}\varsigma_2^\theta(X,A)}_{\text{Log(Variance)}}.
    \end{split}
\end{equation}
It assigns the highest values to data points that are both far from the predicted mean and are associated with a large predicted variance. Essentially, when the model expects being wrong and is. Normal patches where the model is certain and have low error predictions have the lowest anomaly scores. 

\subsection{Model Details}
 While the Patch2Loc task can be applied to different anatomical orientations for the slices, we consider axial slices, such that $A$ is indicates the location of the slice along the vertical dimension (bottom to top of the brain). We select $r=12.5\%$ for the proportion of the axial slice's width and length. Ideally, the smaller the patch, the more precise the resulting heatmap would be. However, in development we found that smaller patch sizes deteriorates Patch2Loc's performance since the patches are more ambiguous. That is, when the patch size is too small, the inherent structural symmetry of the brain causes patches from different locations to appear similar, making it challenging for Patch2Loc to distinguish between them effectively. Contrastingly, larger patches may fail to be useful for identifying anomalies because the model can use the context surrounding an abnormality to accurately predict a location.

For the encoding model $\phi^{}: \mathbb{R}_{\ge0}^{S_1\times S_2}\rightarrow \mathbb{R}^d$, which serves as a common backbone, we use ResNet-18  \citep{he2016deep}, producing a shared latent feature vector with dimension $d=512$.  Since the patch size are small (i.e., $S_1\times S_2=24\times24$), we replace the ResNet-18 first convolution layer with a kernel of size 3 instead of 7.  The latent vector from the backbone $\phi^{}(X)\in\mathbb{R}^d$ is added to a sinusoidal positional encoding of the slice coordinate $A$  \citep{vaswani2017attention}. This representation is input to two separate branches for the mean $\tilde{\mu}^{}:\mathbb{R}^{d}\rightarrow \mathbb{R}^2$ and the log-variance $\tilde{\varsigma}^{}:\mathbb{R}^{d}\rightarrow \mathbb{R}^2$. Each branch consists of four fully-connected  layers with output dimensions $512, 128, 64, 32$, with a batch normalization layer and then a rectified linear unit (ReLU) activation function after each fully-connected layer, followed by a final linear layer to the two-dimensional outputs.    

\subsection{Convolutional Approach for Pixel-level Detection}
\label{sec:conv_approach}
To perform pixel-level abnormality detection, we employ Patch2Loc in a convolutional manner. Specifically,  we feed overlapping patches with a specified stride into the Patch2Loc model, padding patches with zeros when necessary. Subsequently, we utilize the Patch2Loc outputs, predicted mean and variance, along with the corresponding true coordinates to construct the error, variance, and abnormality score map.  This is depicted in the bottom section of Figure~\ref{fig:schematic}. When the stride is 1, the scoring map has the same resolution as the input MRI image.  If a larger stride is used, the abnormality score heatmap will have a lower resolution, but it can be upsampled (e.g., via interpolation) to match the input dimensions. Throughout all experiments, we use a stride of 1.

\section{RELATED WORK}
While not used for unsupervised abnormality segmentation, one prior work by  \cite{NEURIPS2020_d2dc6368} uses self-supervised learning with a task that resembles  Patch2Loc's  location prediction task. The task is a 3D version of a previous context-dependent self-supervised task proposed by \cite{doersch2015unsupervised}. Specifically, the task consists of predicting the discrete relative location of a 3D patch among the possible locations in a $3\times3\times 3$ grid surrounding a center patch that is provided as input. The performance for unsupervised abnormality segmentation was not benchmarked as the task served only as a pretext task to pretrain a backbone network before subsequent supervised segmentation \citep{NEURIPS2020_d2dc6368}. Patch2Loc is distinguished in that its training directly informs the abnormality score. Additionally, Patch2Loc predicts the continuous location of a patch without the need of a center patch as context, using only the slice index as sufficient context for registered brain images. 

Unsupervised abnormality segmentation in neuroimaging has benefited from improving machine learning models. Nonetheless, a notable baseline is based on a simple threshold on the image histogram  \citep{Meissen2022Jul}, which can exceed the performance of baseline methods for certain modalities. Here we describe works relevant to Patch2Loc or those we compare against, a more comprehensive discussion of the related work can be found in Appendix~\ref{sec:related_work}.

Many approaches leverage AE or VAE in novel ways.  Rather than relying on squared error for training, another work by \cite{Meissen2022Aug} uses an AE architecture that reconstructs features obtained from a pretrained encoder using the Structural Similarity Index Measure (SSIM) as the loss function.  SVAE by \cite{Behrendt2022Apr} uses a VAE with transformers to capture the inter-slice dependencies and showed it can improve the results compared to 2D vanilla VAE. The RA method by \cite{Bercea2023Apr} uses an VAE with  a cyclic loss and use the reconstruction error as abnormality score. \cite{baur2021autoencoders} noted that AE and VAE methods suffer from blurry reconstructed images, which hinders their performance. Incorporating a discriminator as in a GAN can improve the reconstruction quality. 

One prior work by \cite{van2021anomaly} used a patch-based auto-encoder with a cycle consistency term and a discriminator to distinguish between the original image and its reconstruction, testing it on specific abnormal tissues. We identified it as the only approach leveraging local features for unsupervised abnormality segmentation in brain MRI. 
However, the method was only applied to a dataset of infarcts (dead brain tissue caused by loss of blood flow) and was not compared for other abnormal tissues.

Another approach is to iteratively restore an image to better match the normal data distribution. PHANEs   \citep{bercea2023reversing} uses a model to restore part of an MRI slice flagged by the RA method  \citep{Bercea2023Apr} to mitigate false positives within the flagged region. This is along the lines of denoising autoencoder (DAE)  \citep{kascenas2022denoising}, which learns to remove correlated noise added to the input images during training. DAE outperformed GAN and VAE approaches, achieving higher Dice score and average precision.

%\subsection{Denoising Diffusion Probabilistic Models}
Denoising diffusion probabilistic models (DDPM)~\citep{ho2020denoising} build on DAEs. One of the first works to use DDPM  for unsupervised abnormality segmentation also proposed learning to denoise simplex noise instead of Gaussian noise to enhance performance~\citep{wyatt2022anoddpm}. While this empirically works, how this matches the fundamental assumptions of diffusion processes is not clear. Specifically, simplex noise is procedurally generated, and is not described by a random process. In contrast, a diffusion model's forward process is described by a Markov chain, with a Markov transition kernel  \citep{sohl2015deep}, typically described by adding Gaussian noise. Nonetheless, the simplex noise version of DDPM formulation essentially defines a training regime for denoising across different signal-to-noise ratios, where longer times correspond to higher noise regimes. During inference, denoising is performed from the initially high noise regime, then new simplex noise is applied at decreasing noise levels, and the denoising process continues. The patched diffusion model (pDDPM) by \citet{behrendt2024patched} adds noise only to a patch of the whole input slice. The rest of the slice gives context for the DDPM to attempt to denoise the noised patch. Different versions of each slice with patches at different locations are used to identify abnormalities within the slice. \citet{behrendt2025guided} used conditional DDPM (cDDPM) to denoise a slice given the latent embedding from a masked auto-encoder (MAE)  \citep{he2022masked} pretrained on normal MRI slices, which is further fine-tuned during training.  
\section{DATA}
We preprocess the neuroimages by sequentially applying the following processes: skull-stripping, registering each to the SRI atlas  \citep{Rohlfing2009Dec}, resampling to a voxel dimension of 1 mm\textsuperscript{3} in the atlas space, applying N4 bias-correction, applying the histogram standardization method proposed by \cite{Nyul2000Feb}, and dividing each MRI image by its 98th percentile. In the penultimate step, the histogram standardization method in  \citep{Nyul2000Feb} uses statistics (e.g., quantiles and the second mode) obtained from a dataset. We use the training dataset to estimate these statistics, and then we standardize every image from all datasets using the same statistics. 

For the training data, we use the IXI dataset  \citep{ixi2015}, which contains MRI scans in both T1- and T2-weighted modalities for 560 subjects.  Following the procedure outlined in  \citet{behrendt2024patched, behrendt2025guided}, a total of 161 samples are set aside for testing, while the remaining data is divided into five sets for cross-validation. Each set consists of 358 training samples and 44 validation samples.

For evaluation, we employ three different datasets:  BraTS21  \citep{menze2014multimodal, baid2021rsna} with 1152 subjects; Multiple Sclerosis Patients with Lesion Segmentation Based on multi-rater consensus (MSLUB)  \citep{lesjak2018novel} with 30 subjects, White Matter Hyperintensity (WMH)  \citep{Kuijf2019Nov} with 60 subjects, and Anatomical Tracings of Lesions After Stroke v2.0 (ATLAS) \citep{Liew2022Jun} with 955 subjects. The first two use T2-weighted and the last two use T1-weighted modality.  These datasets represent different types of abnormal tissues. BraTS21 focuses on brain tumors with varying sizes and convex structures. MSLUB and WMH contain lesions from multiple sclerosis and white matter hyperintensities, respectively, which are  relatively smaller and more scattered than tumors. ATLAS has small convex shaped abnormal tissues. Sample slices from these datasets, along with their ground truth abnormal tissue segmentation, are shown in Figure~\ref{fig:mri_visualization_abnormal}.

\subsection{Patch2Loc Training Details}
\label{sec:patch2loc_training_details}
 For each training batch, we randomly select one slice from each of the 358 training subjects. From the combined pool of 358 slices, we uniformly sample 8096 patches. Patches are discarded if more than 80\% of their pixels are background (i.e., $<$20\% brain tissue). This is depicted in the top section of the schematic diagram shown in Figure~\ref{fig:schematic}, where the yellow outline patch in the top left corner are rejected due to their substantial empty content. This process defines a single batch, and Patch2Loc is trained for 15,000 such batches.
 Adam  \citep{Kingma2014} is used as an optimizer with a learning rate $10^{-2}$ and other hyper-parameters are left as defaults. We use a single NVIDIA V100 GPU to train the model.

\section{RESULTS}
\label{sec:results}
We first compare Patch2Loc with benchmark and state-of-the-art (SOTA) methods  \citep{Meissen2022Aug,Meissen2022Jul,Behrendt2022Apr,Bercea2025Feb,wyatt2022anoddpm,behrendt2024patched,behrendt2025guided} that reflect different methodologies such AE, VAE, GANs, and diffusion models that have published results on these datasets. 

To better understand Patch2Loc's operation, we investigate the distributions of Patch2Loc's location prediction errors and predicted variances, underlying the abnormality score, across normal and abnormal patches. Then we qualitatively illustrate Patch2Loc's abnormality heatmap on representative examples from each dataset.

\begin{table*}[!ht]
\centering
\resizebox{\linewidth}{!}{
\begin{tabular}{lcccccccc}
\hline
 & \multicolumn{2}{c}{BraTS21 (T2)}  & \multicolumn{2}{c}{MSLUB (T2)} & \multicolumn{2}{c}{ATLAS (T1)} & \multicolumn{2}{c}{WMH (T1)} \\
\textbf{Model} & \textbf{$\lceil$DICE$\rceil$ [\%]} & \textbf{AUPRC [\%]} & \textbf{$\lceil$DICE$\rceil$ [\%]} & \textbf{AUPRC [\%]} & \textbf{$\lceil$DICE$\rceil$ [\%]} & \textbf{AUPRC [\%]}& \textbf{$\lceil$DICE$\rceil$ [\%]} & \textbf{AUPRC [\%]} \\
\hline
\textit{Thresh} \citep{Meissen2022Jul} & 30.26  & 20.27  & 7.65  & 4.23 & 4.66 & 1.71  & 10.32 & 4.72\\
\textit{VAE} \citep{baur2021autoencoders} & 33.12 $\pm$ 1.12 & 25.74 $\pm$ 1.37 & 8.10 $\pm$ 0.18 & 4.48 $\pm$ 0.18 & 15.63 $\pm$ 0.73 & 11.44 $\pm$ 0.5 & 7.60 $\pm$ 0.28 & 3.86 $\pm$ 0.40 \\
\textit{SVAE} \citep{Behrendt2022Apr} & 36.43 $\pm$ 0.36 & 30.3 $\pm$ 0.45 & 8.55 $\pm$ 0.11 & 4.8 $\pm$ 0.09 & 10.32 $\pm$ 0.53 & 6.84 $\pm$ 0.44 & 7.18 $\pm$ 0.07 & 2.97 $\pm$ 0.06  \\
\textit{AE} \citep{baur2021autoencoders} & 36.04 $\pm$ 1.73 & 28.8 $\pm$ 1.72 & 9.65 $\pm$ 0.97 & 5.71 $\pm$ 0.80 & 14.04 $\pm$ 0.6 & 10.16 $\pm$ 0.53 & 7.34 $\pm$ 0.08 & 3.43 $\pm$ 0.14 \\
\textit{DAE} \citep{kascenas2022denoising} & 48.82 $\pm$ 3.68 & 49.38 $\pm$ 4.18 & 7.57 $\pm$ 0.61 & 4.47 $\pm$ 0.69 & 15.95 $\pm$ 0.69 & 13.37 $\pm$ 0.62 & 12.02 $\pm$ 1.01 & 8.54 $\pm$ 1.02 \\
\textit{RA} \citep{Bercea2023Apr} & 16.75 $\pm$ 0.51 & 9.98 $\pm$ 0.43 & 3.96 $\pm$ 0.03 & 1.92 $\pm$ 0.04 & 12.21 $\pm$ 0.98 & 8.75 $\pm$ 0.93 & 6.04 $\pm$ 0.45 & 3.15 $\pm$ 0.31 \\
\textit{PHANES} \citep{bercea2023reversing} & 28.42 $\pm$ 0.91 & 21.29 $\pm$ 1.06 & 6.11 $\pm$ 0.27 & 2.98 $\pm$ 0.07 & 17.62 $\pm$ 0.41 & 13.81 $\pm$ 0.48 & 7.55 $\pm$ 0.17 & 3.87 $\pm$ 0.13 \\
\textit{FAE} \citep{Meissen2022Aug} & 44.59 $\pm$ 2.19 & 43.63 $\pm$ 0.47 & 6.85 $\pm$ 0.65 & 3.85 $\pm$ 0.08 & 17.76 $\pm$ 0.16 & 13.91 $\pm$ 0.10 & 8.81 $\pm$ 0.38 & 4.77 $\pm$ 0.26 \\
\textit{DDPM} \citep{wyatt2022anoddpm} & 50.27 $\pm$ 2.67 & 50.61 $\pm$ 2.92 & 9.71 $\pm$ 1.29 & 6.27 $\pm$ 1.58 & 20.18 $\pm$ 0.58 & 17.77 $\pm$ 0.47 & 12.06 $\pm$ 0.97 & 8.89 $\pm$ 0.89 \\
\textit{pDDPM} \citep{behrendt2024patched} & 53.61 $\pm$ 0.51 & 55.08 $\pm$ 0.54 & 12.83 $\pm$ 0.40 & \textit{10.02 $\pm$ 0.36} & 19.92 $\pm$ 0.24 & 17.84 $\pm$ 0.10 & 10.13 $\pm$ 0.53 & 7.52 $\pm$ 0.56 \\
cDDPM \citep{behrendt2025guided} & \textit{56.30 $\pm$ 1.25} & \textbf{58.82 $\pm$ 1.56} & \textit{14.04 $\pm$ 1.16} & \textbf{10.97 $\pm$ 1.17} & \textit{24.22 $\pm$ 1.10} & \textbf{22.22 $\pm$ 1.15} & \textit{11.59 $\pm$ 0.93} & \textit{9.26 $\pm$ 1.07} \\
\textit{Patch2Loc} (Ours) & \textbf{59.50 $\pm$ 1.45} & \textit{55.40 $\pm$ 1.30} & \textbf{14.30 $\pm$ 1.40} & 8.70 $\pm$ 1.20 & \textbf{25.50 $\pm$ 0.73} & \textit{22.00 $\pm$ 1.70} & \textbf{15.70 $\pm$ 1.70} & \textbf{10.10 $\pm$ 1.70} \\
\hline
\end{tabular}
}
\caption{Comparison of the evaluated models with the best results highlighted in bold, and second best italicized. For all metrics, the mean $\pm$ standard deviation across the different folds are reported.}
\label{tab:benchmarks}
\end{table*}

\subsection{Quantitative Results for Unsupervised Abnormality Segmentation }
Performance is measured, following previous work  \citep{behrendt2024patched, behrendt2025guided}, in terms of both average precision (area under the precision-recall curve) and the best possible Dice-coefficient (highest possible F1 score) $\lceil\text{Dice}\rceil$  per subject and then averaged over the BraTS, MSLUB, ATLAS, and WMH datasets. The results are in Table~\ref{tab:benchmarks}.  Patch2Loc often matches or outperforms the best performing method, with a wide margin for $\lceil{\text{Dice}}\rceil$ on the WMH dataset.

\subsection{Abnormality Score Analysis}
\begin{figure}[tbh]
    \centering
        \includegraphics[width=\columnwidth]{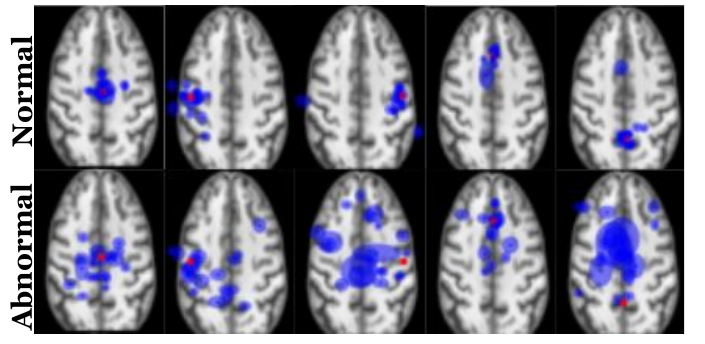}
    \caption{Visualization of Patch2Loc's output (blue ellipses) for patches captured from the same location (red dot) across different subjects in the BraTS dataset, overlaid on a representative  T1-weighted registered slice (without abnormalities).  (Top row) Predictions of normal patches. (Bottom row) Corresponding predictions of abnormal patches. The predicted Gaussian distribution is visualized as an ellipse, where the center represents the predicted mean, and the major and minor axes correspond to two standard deviations.}
\label{fig:distributions}
\end{figure}

We illustrate the operation of Patch2Loc by examining the predicted distribution (mean and variance) of patches extracted from the same spatial location across subjects. For patch-level analysis, we obtain a set of non-overlapping patches from the abnormal datasets across all slices and individuals. We categorize a patch as either normal if it contains less than 10\% of abnormal tissues or abnormal if it comprises over 90\% of abnormal tissues. Patches falling within the 10\% to 90\% abnormal tissue percentage range are separately analyzed.  We compare Patch2Loc's predictions for normal and abnormal patches drawn from the BraTS dataset in Figure~\ref{fig:distributions}. For normal patches, we observe low uncertainty (i.e., smaller ellipses) and low error, as the predictions are tightly clustered around the true location. Conversely, for abnormal patches, the model exhibits higher uncertainty (i.e., larger ellipses) and greater prediction error, with the predicted means deviating significantly from the true location.

\begin{figure*}[hbt]
\centering
\includegraphics[width=\textwidth]{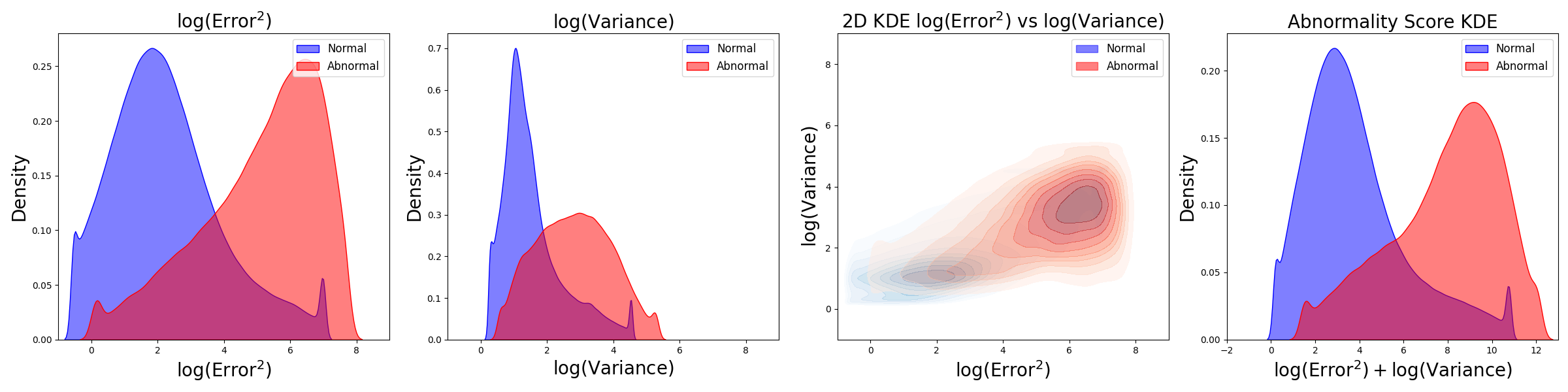}
\caption{From left to right: 1D KDE for log-error, and log-variance, 2D KDE for log-error and log-variance, and 1D KDE for the abnormality score for normal patches (blue) and abnormal patches (red) for the BraTS dataset.}
\label{fig:kde_brats}
\end{figure*}

Figure~\ref{fig:kde_brats} shows kernel density estimates (KDEs) of Patch2Loc’s location prediction log-errors and predicted log-variance on normal and abnormal patches. (Figure~\ref{fig:kde_all} in the Appendix shows the results for all datasets).
There is a clear separation between the normal and abnormal patches in the space of log-error and log-variance although there is some overlap. This overlap occurs because Patch2Loc accurately predicts the location of abnormal patches with low variance. Such predictions are more likely when the abnormal patches are located at the edges of the brain, where Patch2Loc can utilize the surrounding empty space to make precise location predictions. This is mitigated by having overlapped patches as explained in Section~\ref{sec:conv_approach}. 

To investigate Patch2Loc on the partially abnormal patches (i.e., the patches that have abnormal content between 10\% and 90\%), we calculate the Spearman correlation between the abnormal content and log-error$^2$, log-variance, and the abnormal score (i.e., their sum) for each dataset as shown in Table~\ref{tab:spearman_corr}. In all datasets, the abnormality score of the sum better correlates with the level of proportion of abnormal tissue in a patch compared to either log-error or log-variance alone, highlighting their complementary information.  
\begingroup
\setlength{\tabcolsep}{3.0pt} % Default 
\begin{table}[htb]
\centering
\caption{Spearman correlation (\%) between a patch's portion of abnormality and the log-error, log-variance, or abnormality score.}
\label{tab:spearman_corr}
{\footnotesize

\begin{tabular}{lcccc}
\hline
\textbf{Metric} & \textbf{BraTS} & \textbf{MSLUB} & \textbf{ATLAS} & \textbf{WMH} \\
\hline
\textbf{Log-error} & 37 & 15 & 20 & 12 \\
\textbf{Log-variance} & 34 & 14 & 18 & 11 \\
\textbf{Score} & 40 & 17 & 23 & 13 \\
\hline
\end{tabular}
}
\end{table}
\endgroup 

\subsection{Qualitative Analysis}
\begin{figure}[htb]
    \includegraphics[width=\columnwidth,trim=0 30 0 27, clip]{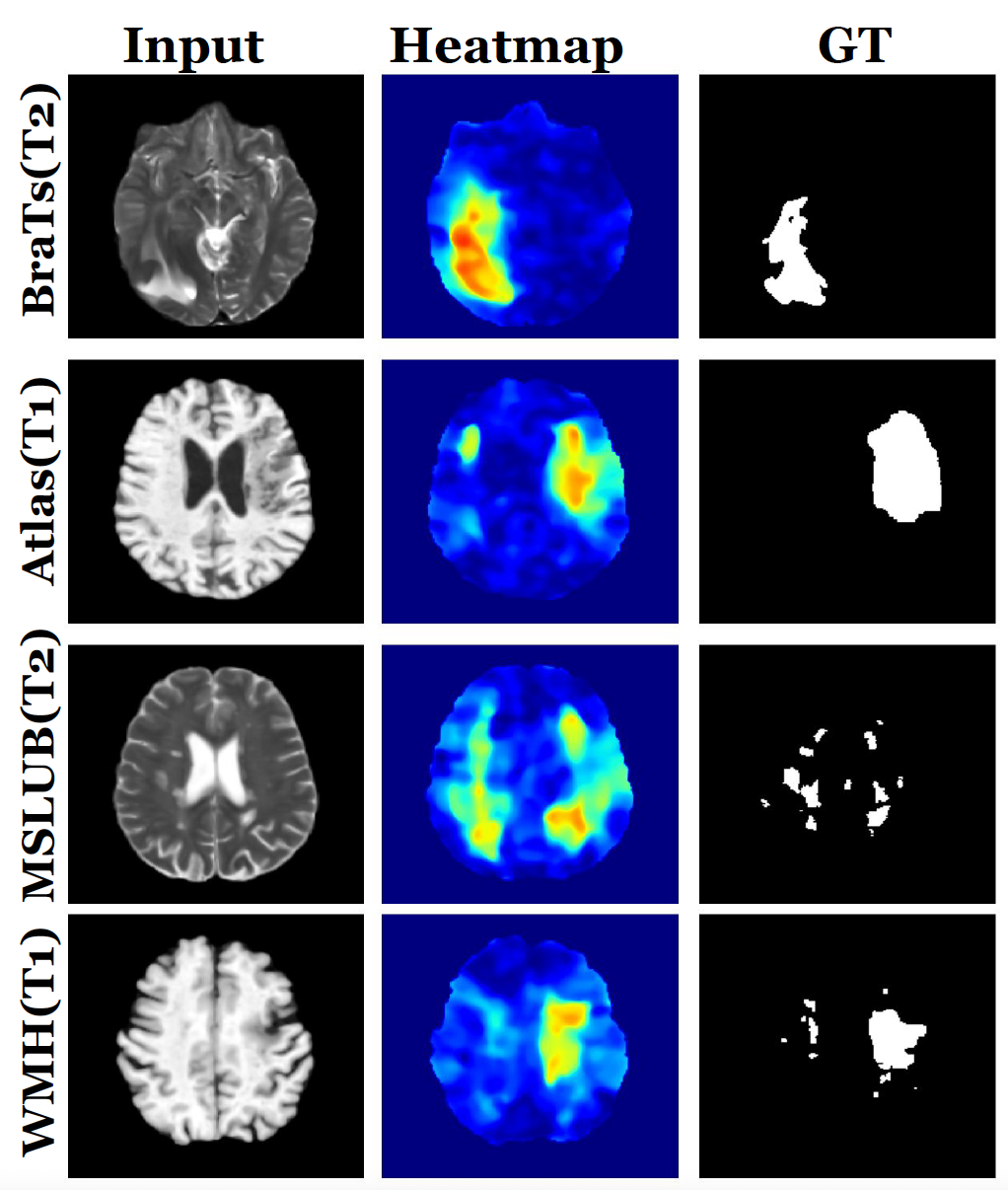}
    \caption{Visualization of our model's anomaly detection performance on pathological slices from the BraTS, ATLAS, MSLUB, and WMH datasets, the model's heatmap successfully localizes abnormalities, showing high correlation with the ground truth masks (white). Colormap ranges from 0 (blue) to 12 (red).}
    \label{fig:mri_visualization_abnormal}
\end{figure}

In Figure~\ref{fig:mri_visualization_abnormal}, we present visualizations of MRI slices from the BraTS, ATLAS, MSLUB and WMH datasets, along with the corresponding heatmaps. While only one representative slice from each dataset are shown, they reflect a consistent pattern observed across both datasets. Additional visualizations are provided in the Appendix (Figure~\ref{fig:appendix_images_brats} for BraTS, Figure~\ref{fig:appendix_images_mslub} for MSLUB, Figure~\ref{fig:appendix_images_atlas} for ATLAS, and Figure~\ref{fig:appendix_images_wmh} for WMH). 

We also provide visualizations for slices from IXI test set (i.e. normal subjects) where it shows low abnormal scores in Figure~\ref{fig:normal_controls} with more slices in Figure~\ref{fig:appendix_images_ixi} in the Appendix. For fair comparison, the colormap is fixed to [0, 12], as the histograms show scores are typically concentrated within this range, even though lower/higher values may occur.
\begin{figure}[htb]
\centering\includegraphics[width=0.475\columnwidth,trim=70 110 40 55, clip]{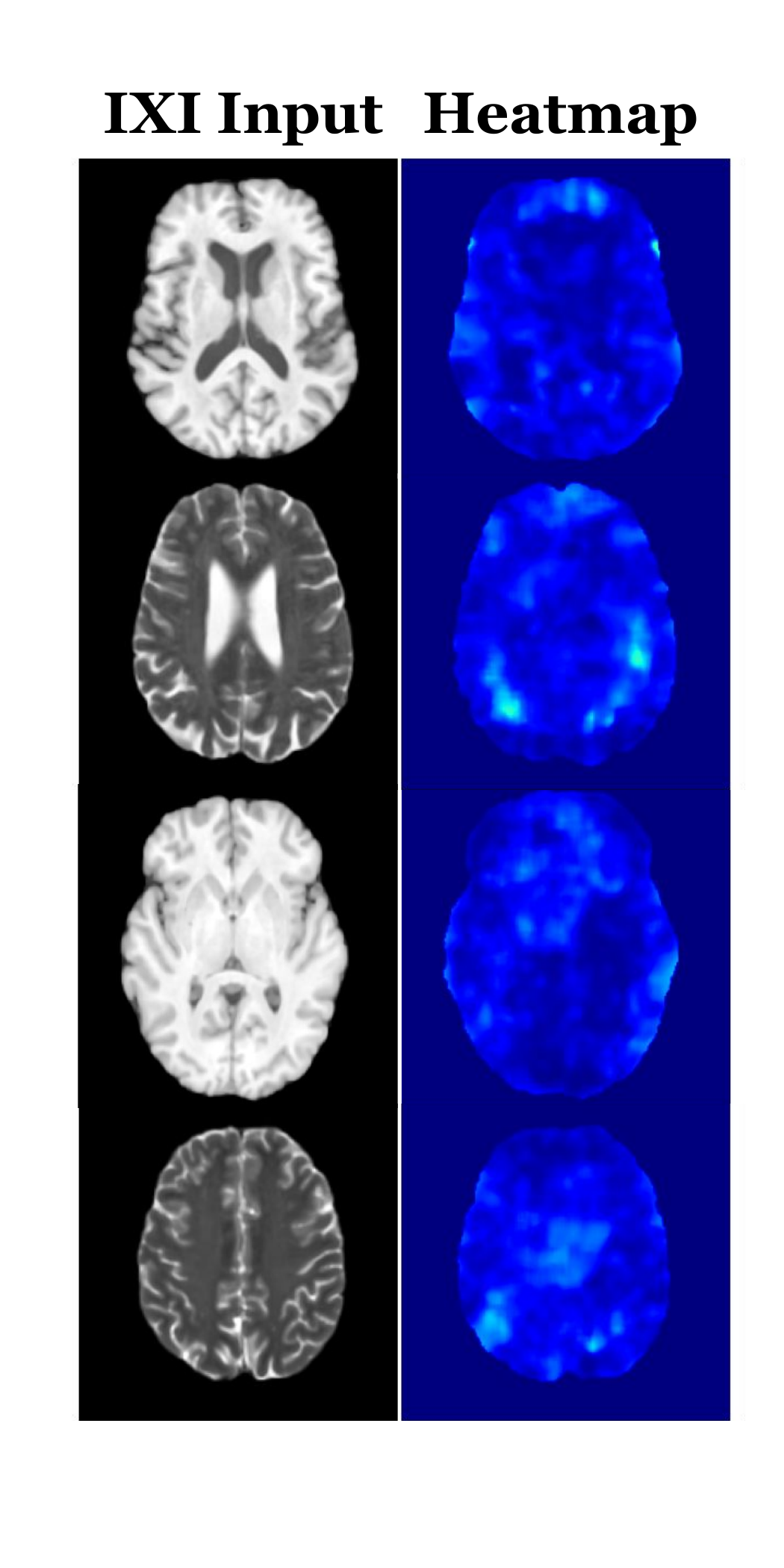}  
    \caption{Visualization of our model's heatmap on healthy control slices (T1, T2, T1, T2, across the rows) from the IXI dataset testing set.  The model correctly produces low, diffuse activations. Colormap ranges from 0 (blue) to 12 (red).}
    \label{fig:normal_controls}
\end{figure}

From these visualizations, we observe that the heatmaps spatially correspond to the ground truth abnormality regions. The correspondence for the BraTS is apparent. The abnormalities in BraTS are tumors, and patches from overlapping tumors will lack the normal anatomical structure necessary for accurate location prediction. In the case of MSLUB, the presence of normal tissue surrounding small and scattered abnormal regions provides contextual cues that may allow Patch2Loc to have correct location predictions. The same applies to WMH where the long but narrow abnormal structure also provide contextual cues. Likewise, in ATLAS, if the abnormal region is very small (e.g., in the third row in Figure~\ref{fig:appendix_images_atlas}), it will go unpredicted.  However, the presence of  abnormalities within a patch may still cause the location prediction or increased variance in the prediction, such that Patch2Loc's abnormality score does correlate with these smaller abnormalities. That is, Patch2Loc's abnormality score is high due to in-distribution ambiguity, with large variance estimates, or out-of-distribution patches causing error in the location prediction, and a wide range of variance estimates. These different cases can be see in the KDE plots in Figure~\ref{fig:kde_all}.

\section{DISCUSSION}
The results showcase that Patch2Loc  advances the state of the art for unsupervised abnormality segmentation  in neuroimages through an intuitive approach.  Its promising performance meets or exceeds the best performing benchmark cDDPM  \citep{behrendt2025guided}, which has a more complicated structure for training and inference due the combination of the masking autoencoder and DDPM model. In contrast, Patch2Loc is unique in terms of its dependence of local features, without global context, and it does not depend on reconstruction errors compared to the others. During test time, our method does not depend on any hyperparameter such as the amount or type of noise added.  For instance, DAE  \citep{kascenas2022denoising} has to search for the correlated noise parameter that gives the best performance. Generally, prior work requires extensive hyperparameter search during training (especially for methods involving GANs) and/or testing. In contrast, Patch2Loc's hyper-parameters are optimized in terms of the prediction task solely for the normal anatomical structure of brain images from the IXI brain slices before the models (one for T1 and T2) are evaluated on different datasets. 

One limitation of Patch2Loc is its ability to infer the correct location when abnormalities are smaller than the patch size. Smaller patches deteriorate prediction performance on normal patches, due to the high similarity of patches from widely different locations.  We note that using a variational autoencoder for the patch with location information as additional input could lead to a complementary framework to Patch2Loc, where the patch reconstruction error and uncertainty regarding the latent embedding could augment the abnormality score.

\section{CONCLUSION}
\label{sec:conclusion}
We have introduced Patch2Loc, a novel self-supervised learning task and model tailored for lesion detection in neuroimages, and demonstrated its effectiveness for unsupervised abnormal tissue segmentation. Our approach leverages the regularities of location-specific image features to identify abnormalities in brain tissues and directly incorporates uncertainty estimates using the $\beta$-NLL framework  \citep{seitzer2022on}. Unlike prior methods that focus on global features, Patch2Loc emphasizes local representations, enhancing its applicability to unsupervised abnormality segmentation in brain MRI. Our method does not need any hyperparameter adjustment during the inference time such as the level of added noise as required by state-of-the-art methods  \citep{behrendt2025guided,behrendt2024patched} that rely on denoising diffusion models.  This work introduces a new perspective on unsupervised abnormality segmentation  in neuroimaging and lays the foundation for future research in this direction. 

% \textbf{Large Language Models}: LLM are used to polish and improve the writing of this manuscript. 

\section*{Acknowledgements}
Research was carried out with the support of the University of Delaware Research Foundation. This research was supported in part through the use of Information Technologies (IT) resources at the University of Delaware, specifically the high-performance computing resources. The authors would like to thank Heidi Kecskemethy and Rahul Nikam from Nemours Children's Hospitial, Sokratis Makrogiannis from Delaware State University, and Curtis Johnson from the University of Delaware for engaging discussions regarding computer-assisted neuroradiology. We thank Finn Behrendt for providing his code as our implementation is based on and extends the publicly available code of \cite{behrendt2025guided}.

\bibliography{references}
\bibliographystyle{apalike}
\section*{Checklist}

\begin{enumerate}

  \item For all models and algorithms presented, check if you include:
  \begin{enumerate}
    \item A clear description of the mathematical setting, assumptions, algorithm, and/or model. [Yes]
    \item An analysis of the properties and complexity (time, space, sample size) of any algorithm. [Not Applicable]
  \end{enumerate}

  \item For any theoretical claim, check if you include:
  \begin{enumerate}
    \item Statements of the full set of assumptions of all theoretical results. [Not Applicable]
    \item Complete proofs of all theoretical results. [Not Applicable]
    \item Clear explanations of any assumptions. [Yes]     
  \end{enumerate}

  \item For all figures and tables that present empirical results, check if you include:
  \begin{enumerate}
    \item The code, data, and instructions needed to reproduce the main experimental results (either in the supplemental material or as a URL). [Yes]
    \item All the training details (e.g., data splits, hyperparameters, how they were chosen). [Yes]
    \item A clear definition of the specific measure or statistics and error bars (e.g., with respect to the random seed after running experiments multiple times). [Yes]
    \item A description of the computing infrastructure used. (e.g., type of GPUs, internal cluster, or cloud provider). [Yes]
  \end{enumerate}

  \item If you are using existing assets (e.g., code, data, models) or curating/releasing new assets, check if you include:
  \begin{enumerate}
    \item Citations of the creator If your work uses existing assets. [Yes]
    \item The license information of the assets, if applicable. [Yes]
    \item New assets either in the supplemental material or as a URL, if applicable. [No]
    \item Information about consent from data providers/curators. [Not Applicable]
    \item Discussion of sensible content if applicable, e.g., personally identifiable information or offensive content. [Not Applicable]
  \end{enumerate}

  \item If you used crowdsourcing or conducted research with human subjects, check if you include:
  \begin{enumerate}
    \item The full text of instructions given to participants and screenshots. [Not Applicable]
    \item Descriptions of potential participant risks, with links to Institutional Review Board (IRB) approvals if applicable. [Not Applicable]
    \item The estimated hourly wage paid to participants and the total amount spent on participant compensation. [Not Applicable]
  \end{enumerate}

\end{enumerate}

\clearpage
\appendix
\thispagestyle{empty}

% Supplementary material: To improve readability, you must use a single-column format for the supplementary material.
\onecolumn

\section{RELATED WORK}
\label{sec:related_work}
\label{related_work_first_category}
Unsupervised abnormality segmentation in neuroimaging has benefited from improving machine learning models including autoencoders (AE), variational autoencoders (VAE)  \citep{baur2021autoencoders}, the use of adversarial losses involving discriminators as in generative adversarial neural networks (GANs) \citep{goodfellow2014generative}, transformers~\citep{vaswani2017attention}, and diffusion models~\citep{ho2020denoising}.

Many approaches leverage AE or VAE in novel ways. An early work by \cite{zimmerer2018context} introduced \emph{ceVAE}, which  combines variational and context autoencoders to compute abnormality scores from density and reconstruction error. Another work by \cite{sato2019predictable} uses a VAE with the abnormality score defined as the reconstruction error divided by the estimated variance. Rather than relying on squared error for training, another work by \cite{Meissen2022Aug} uses an AE architecture that reconstructs features obtained from a pretrained encoder using the Structural Similarity Index Measure (SSIM) as the loss function.  SVAE by \cite{Behrendt2022Apr} uses a VAE with transformers to capture the inter-slice dependencies and showed it can improve the results compared to 2D vanilla VAE. The RA method by \cite{Bercea2023Apr} uses an VAE with  a cyclic loss and use the reconstruction error as abnormality score.

Previous studies \citep{baur2021autoencoders,pinaya2022unsupervised} noted that AE and VAE methods suffer from blurry reconstructed images, which hinders their performance. Incorporating a discriminator as in a GAN can improve the reconstruction quality. However, methods incorporating GAN-losses suffer from training instability due the competition between the discriminator and decoder  \citep{brock2018large}. Nonetheless, the work by \cite{chen2018unsupervised} proposed an adversarial autoencoder (AAE) that enforces a prior on the latent space and a cyclic objective for encoder consistency such that an image and its reconstruction are similar in the latent space. 
% Not compared
AnoVAGAN by \cite{baur2019deep} is a variational autoencoder with spatially organized latent codes and a GAN loss. 

Another approach is to iteratively restore an image to better match the normal data distribution, using the number of restoration steps as an abnormality score  \citep{chen2020unsupervised}. PHANEs by  \cite{bercea2023reversing} uses a model to restore part of an MRI slice flagged by the RA method  \citep{Bercea2023Apr} to mitigate false positives within the flagged region. 

One prior work by \cite{van2021anomaly} used a patch-based auto-encoder with a cycle consistency term and a discriminator to distinguish between the original image and its reconstruction, testing it on specific abnormal tissues. We identified it as the only approach leveraging local features for unsupervised abnormality segmentation in brain MRI. However, it suffers from instability due to adversarial training and difficulty in balancing multiple loss terms, which is challenging in unsupervised setting. The method was applied to detect infarcts, but it was not applied to abnormality segmentation and was not compared against any existing work.  %Thus, it is questionable if patch-level reconstruction would be competitive compared to slice-level reconstructions. 

While the aforementioned works advanced unsupervised abnormality detection for neuroimages, all are outperformed by a significant margin by using a denoising autoencoder (DAE)  \citep{kascenas2022denoising}, which learns to remove correlated noise added to the input images during training. DAE outperformed GAN and VAE approaches, achieving higher Dice score and average precision.

A promising work by \cite{pinaya2022unsupervised} introduced a combined vector-quantized variational autoencoder to learn spatial latent representation for brain slices and then used it to train an autoregressive transformer that operates on patches of the latent representation using different raster orders. Unfortunately, the results are not comparable as they are limited to the FLAIR modality, with 15,000 normal scans sourced from the UK Biobank (UKB) dataset, which is not freely available, and the method was not tested with other modalities or datatsets. It should be noted that the IXI dataset used for training in our benchmark does not have FLAIR modality.

While not used for unsupervised abnormality segmentation, one prior work by  \cite{NEURIPS2020_d2dc6368} uses self-supervised learning with a task that resembles  Patch2Loc's  location prediction task. It is a 3D version of a previous context-dependent self-supervised task of predicting a patch's relative location with respect to a context patch as a classification problem  \citep{doersch2015unsupervised}. Specifically, the task consists of predicting the discrete relative location of a 3D patch among the possible locations in a $3\times3\times 3$ grid surrounding a center patch that is provided as input. Essentially, the goal is to infer the relative location of one patch with respect to center patch as context. The performance for unsupervised abnormality segmentation was not benchmarked as the task served only as a pretext task to learn latent representation via a backbone network that improves subsequent supervised segmentation performance  \citep{NEURIPS2020_d2dc6368}. Patch2Loc is distinguished in that its training directly informs the abnormality score. Additionally, Patch2Loc predicts the continuous location of a patch without the need of a center patch as context, using only the slice index as sufficient context for registered brain images.

Many of the top performing methods for unsupervised abnormality detection use denoising diffusion probabilistic models (DDPM). Unless stated otherwise, all methods use the absolute reconstruction error between input and denoised ones as the abnormality score. One of the first works to use DDPM  \citep{wyatt2022anoddpm}  also proposed learning to denoise simplex noise instead of Gaussian noise to enhance performance. While this empirically works, how this matches the fundamental assumptions of diffusion processes is not clear. Specifically, simplex noise is procedurally generated, and is not described by a random process. In contrast, a diffusion model's forward process is described by a Markov chain, with a Markov transition kernel  \citep{sohl2015deep}, typically described by adding Gaussian noise, but the Markov chain can also be described by a Bernoulli transition kernel  \citep{sohl2015deep}. Without a known Markov transition kernel, the derivation of the DDPM formulation  \citep{ho2020denoising} may not be applicable to simplex noise, since the Gaussian assumption is exploited to define the forward process posterior mean. Nonetheless, ignoring the validity, the DDPM formulation essentially defines a training regime for denoising across different signal-noise levels, where longer times correspond to higher noise regimes. During inference, denoising is performed from the initially high noise regime, then new simplex noise is applied at decreasing noise levels, and the process is repeated.

Another work by \cite{pinaya2022fast} trained diffusion models on the learned spatial latent features and during inference time the abnormal features are inpainted with normal ones. Then the reconstructed image is obtained using the denoised spatial latent features that are fed to a decoder. Although their work showed an impressive performance for head CT, for brain MRI, their work did not exceed their earlier work  \citep{pinaya2022unsupervised}. This was extended by another work by \cite{liang2023modality} that leveraged a diffusion model for cyclic translations between different modalities and implemented a conditional model, akin to a restoration-based approach. While the model shows superior performance compared to all other techniques mentioned above it requires different modalities. 

Similarly, another work by \cite{bercea2023mask} introduced an iterative inpainting technique to address the noise paradox in order to mitigate the false positives in a high noise regime. The approach initially estimates mask for abnormal tissues based on reconstruction error in the high noise regime. Then an iterative method is applied to inpaint regions of the mask using information outside the mask in a lower noise regime.  

The patched diffusion model (pDDPM) by \cite{behrendt2024patched} performs the noising and denoising within a patch of the whole slice. That is the rest of the slice gives context for the DDPM of the noised patch. Versions of each slice with patches at different locations are used to identify abnormalities within the slice. In a follow-up work \citep{behrendt2025guided} a conditional DDPM (cDDPM) is used for denoising. The conditioning signal is the latent embedding from a masked auto-encoder (MAE)  \citep{he2022masked} pretrained on normal MRI slices, which is further fine-tuned during training. This conditioning gives the model global perspective of structure while performing the denoising. 

\section{ADDITIONAL FIGURES}
\begin{figure*}[hbtp]
\centering
\includegraphics[width=1\textwidth]{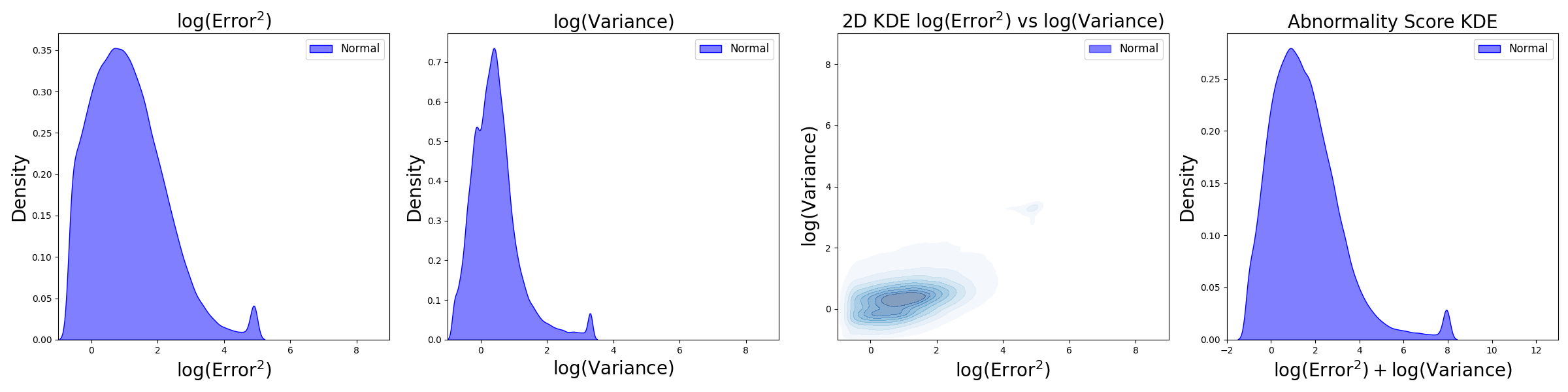}
\includegraphics[width=1\textwidth]{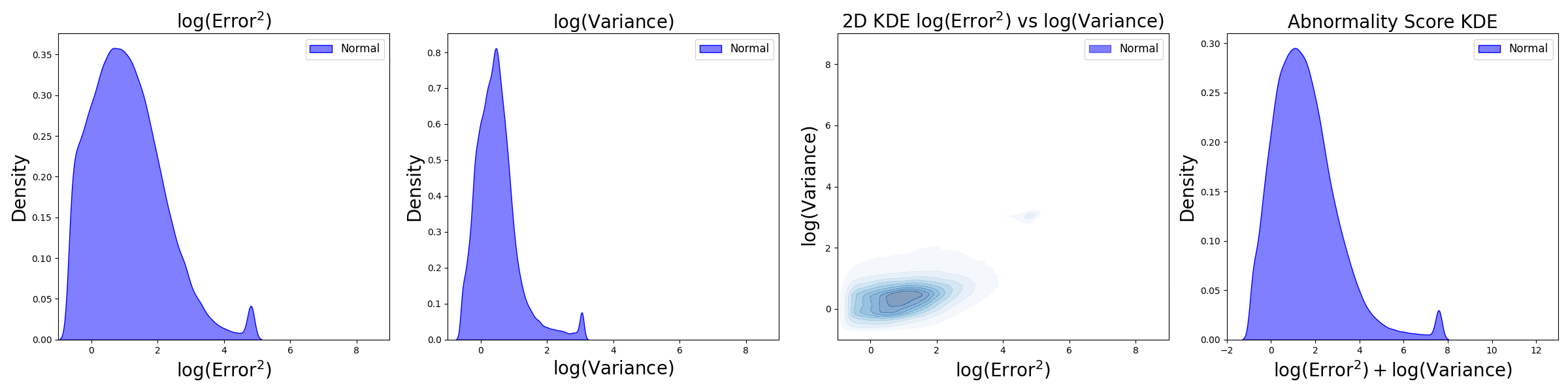}
\includegraphics[width=1\textwidth]{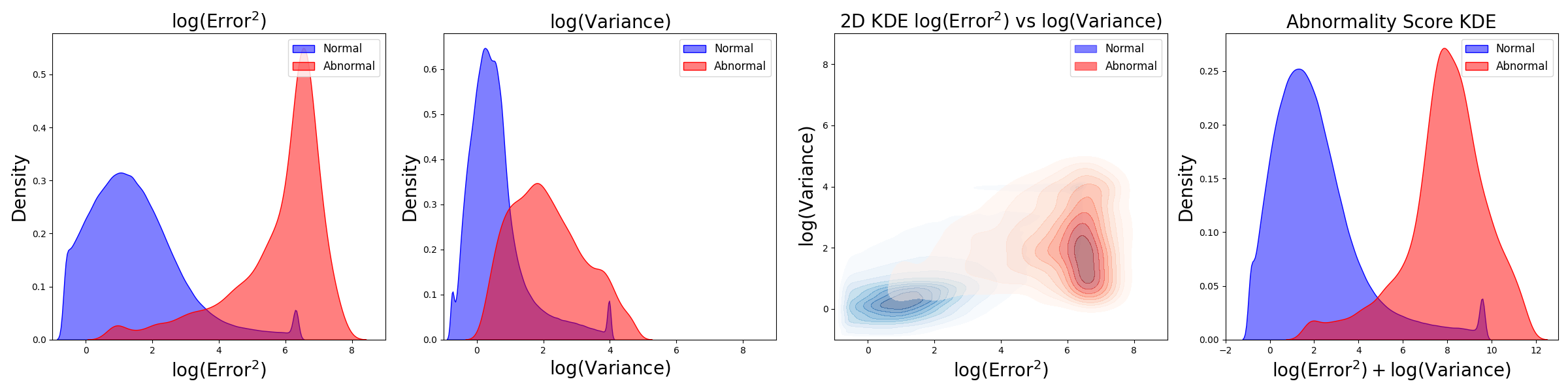}
\includegraphics[width=1\textwidth]{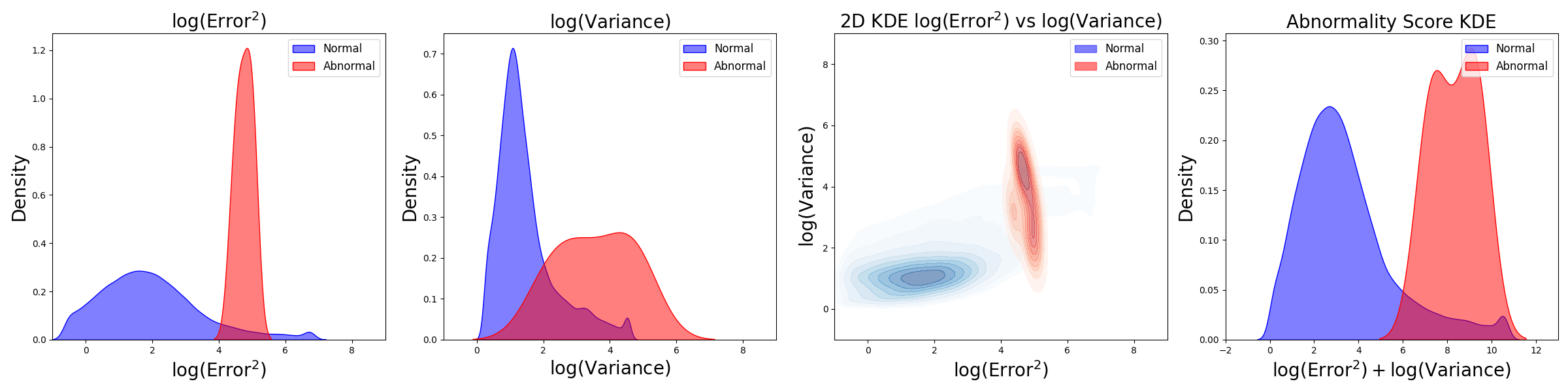}
\includegraphics[width=1\textwidth]{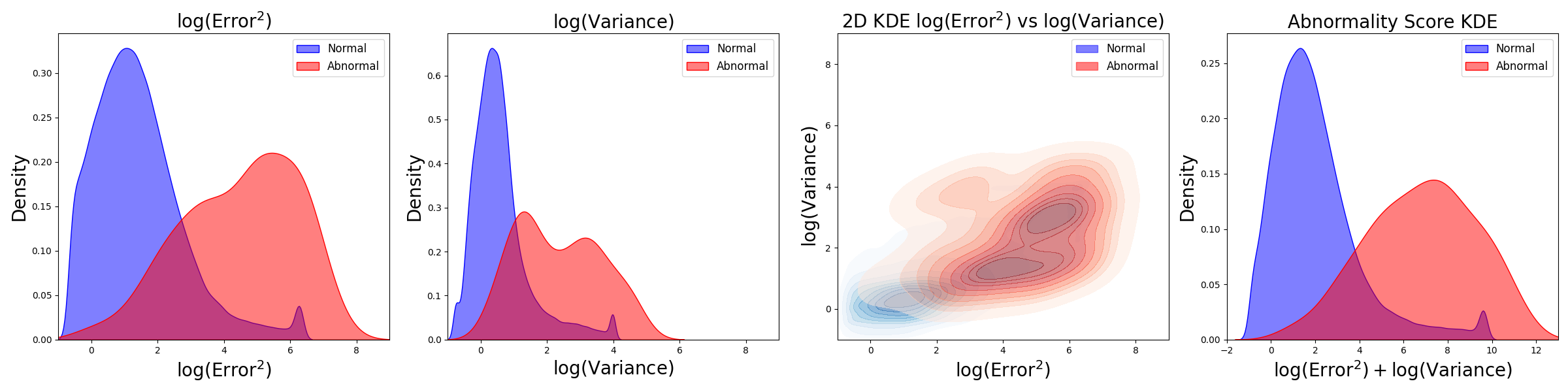}
\caption{From left to right: 1D KDE for log-error, and log-variance, 2D KDE for log-error and log-variance, and 1D KDE for the abnormality score for normal patches (blue) and abnormal patches (red). Top to bottom: test set IXI (T1), test set IXI (T2), ATLAS (T1), MSLUB (T2), and WMH (T1) datasets.}
\label{fig:kde_all}
\end{figure*}

\begin{figure*}[btph]
\centering
\includegraphics[width=0.6125\linewidth,trim=55 180 50 40, clip]{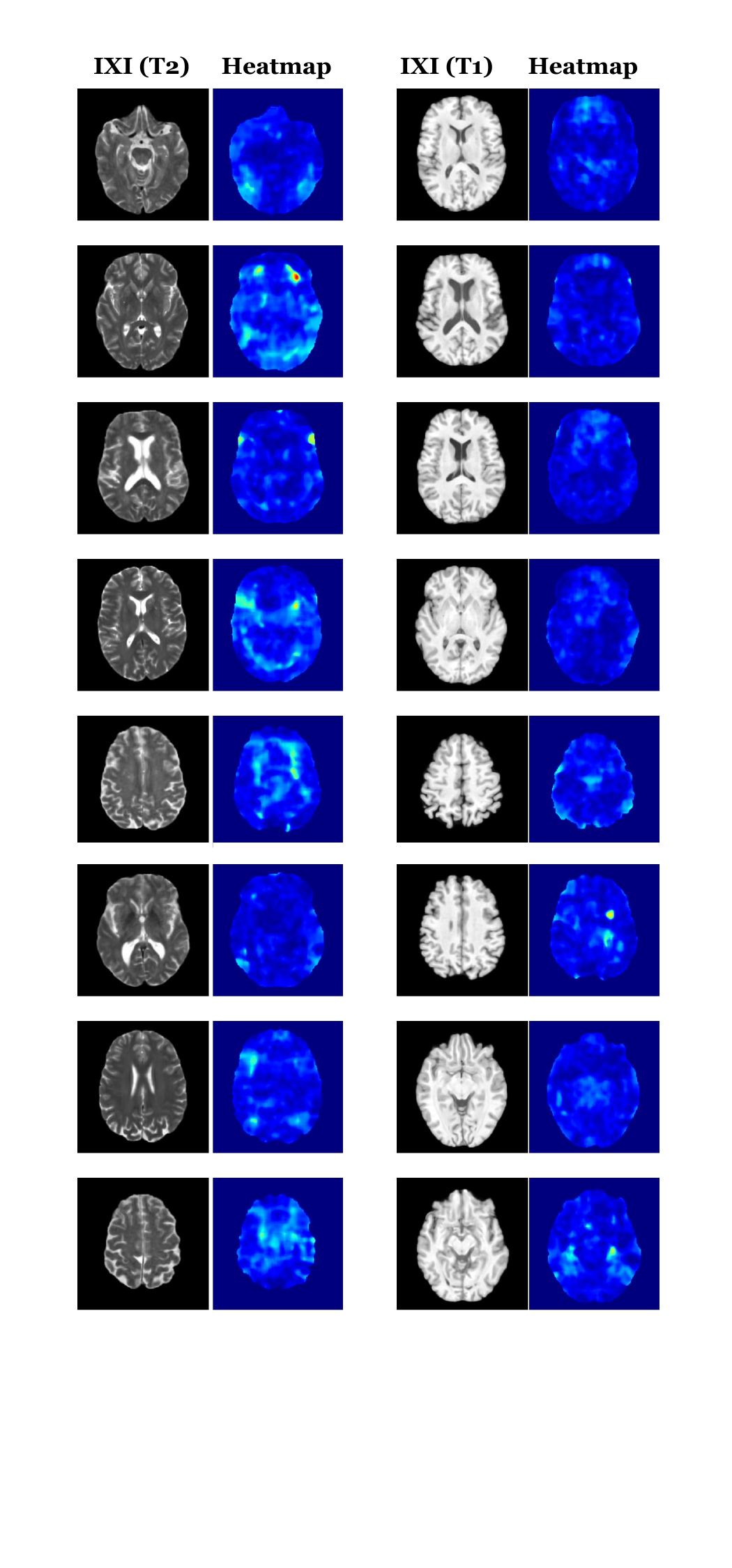}
\caption{Visualization of slices from IXI (T2 and T1). Colormap ranges from 0 (blue) to 12 (red).}
\label{fig:appendix_images_ixi}
\end{figure*}

\begin{figure}[t]
\centering
\includegraphics[width=\columnwidth,trim=80 275 70 50, clip, clip]{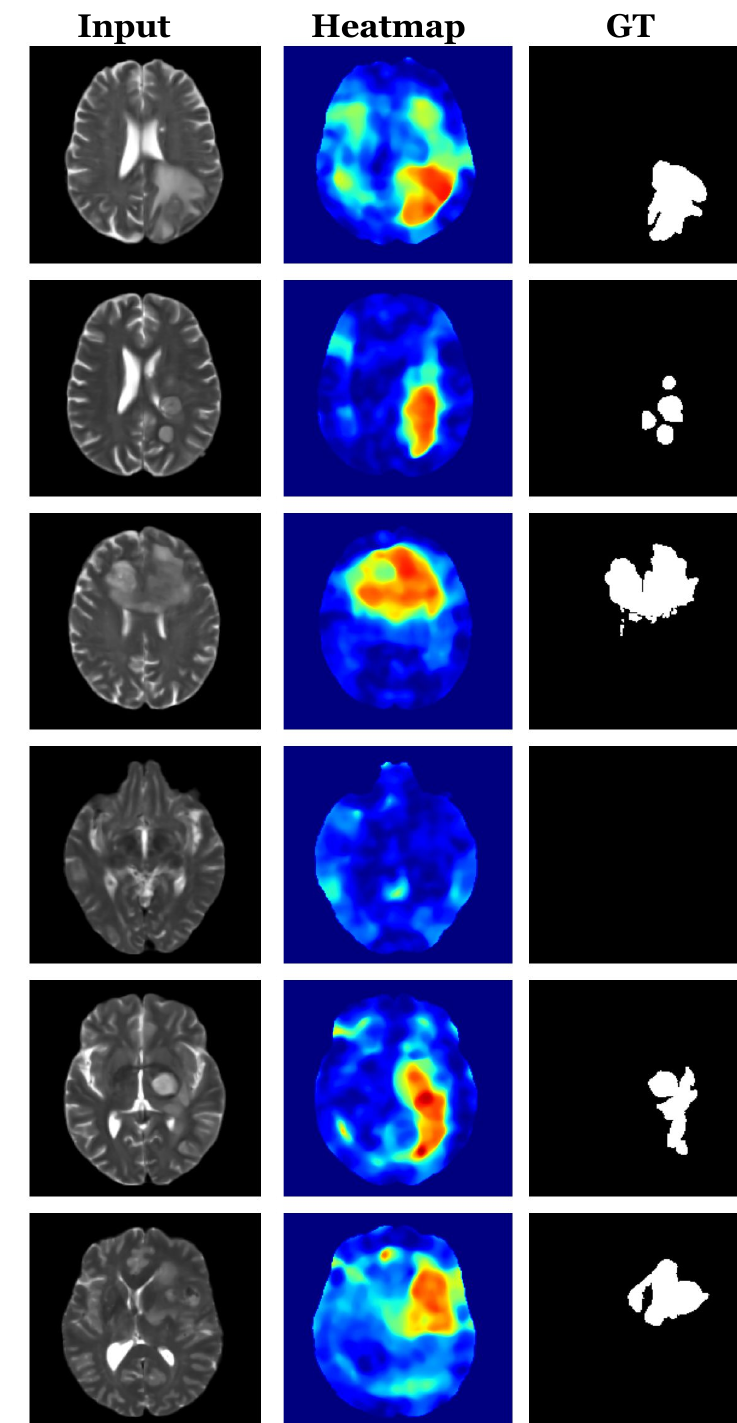}
\caption{Visualization of slices from BraTS (T2). From left to right: the input slice, the heatmap (ranges from 0 (blue) to 12 (red)), and the ground truth for abnormal tissues.}
\label{fig:appendix_images_brats}
\end{figure}

\begin{figure}[t]
\centering
\includegraphics[width=\columnwidth,trim=80 290 70 45, clip]{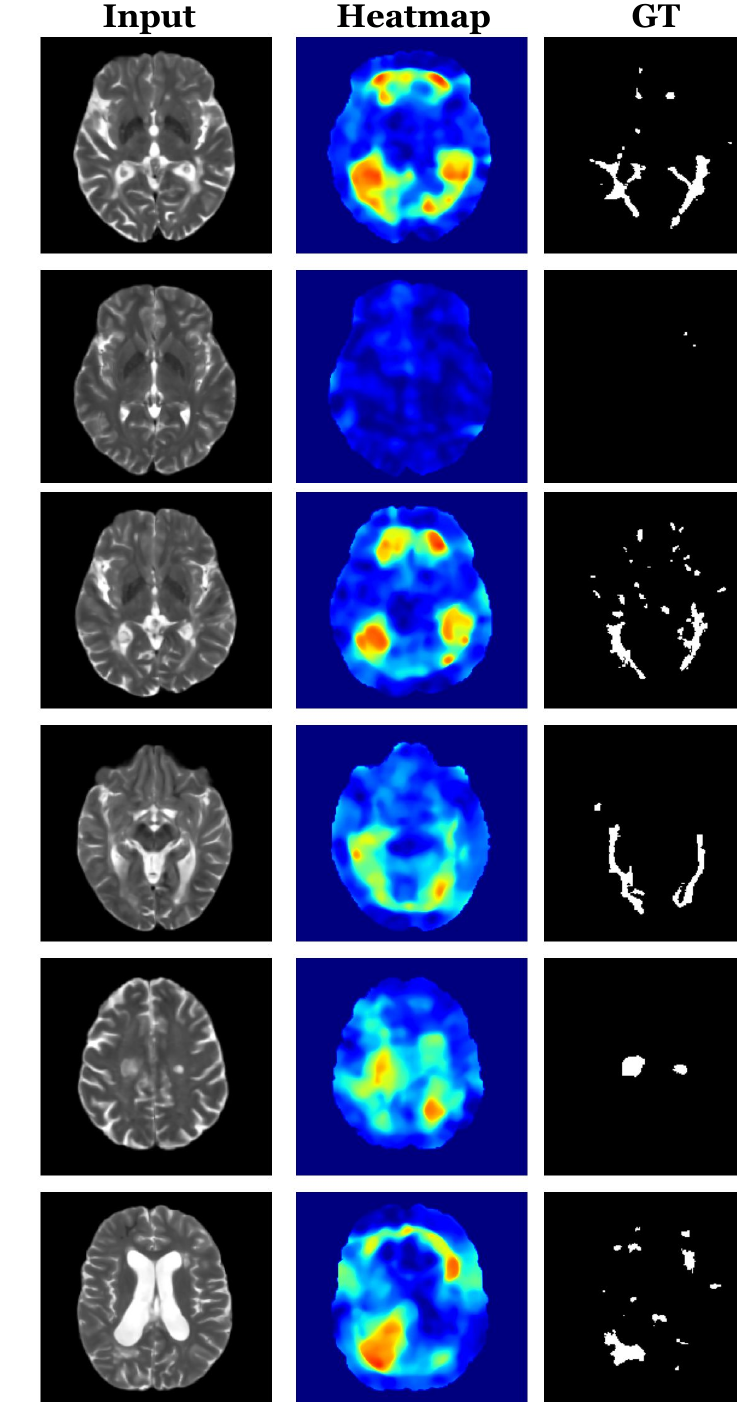}
\caption{Visualization of slices from MSLUB (T2). From left to right: the input slice, the heatmap (ranges from 0 (blue) to 12 (red)), and the ground truth for abnormal tissues.}
\label{fig:appendix_images_mslub}
\end{figure}

\begin{figure}[t]
\centering
\includegraphics[width=\columnwidth,trim=88 300 66 52, clip]{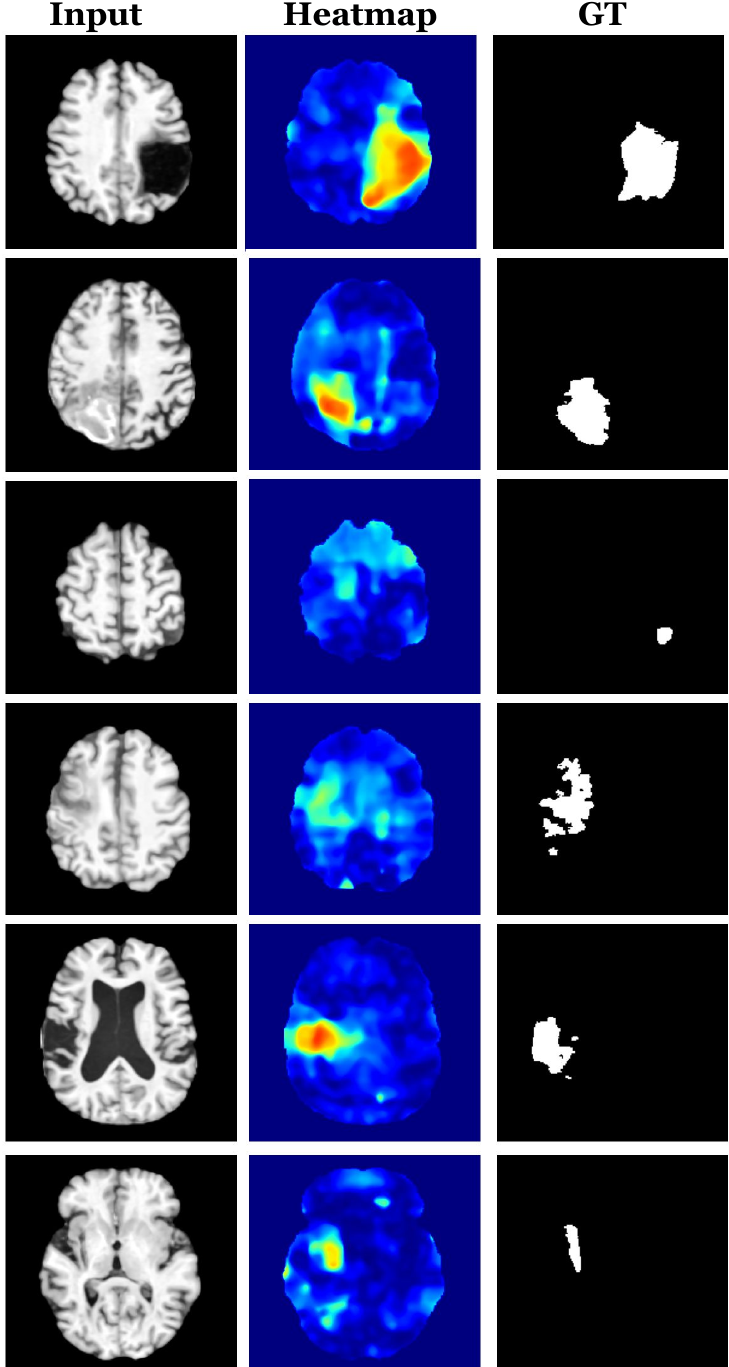}
\caption{Visualization of slices from ATLAS  (T1). From left to right: the input slice, the heatmap (ranges from 0 (blue) to 12 (red)), and the ground truth for abnormal tissues.}
\label{fig:appendix_images_atlas}
\end{figure}

\begin{figure}[t]
\centering
\includegraphics[width=\columnwidth,trim=80 210 80 118, clip]{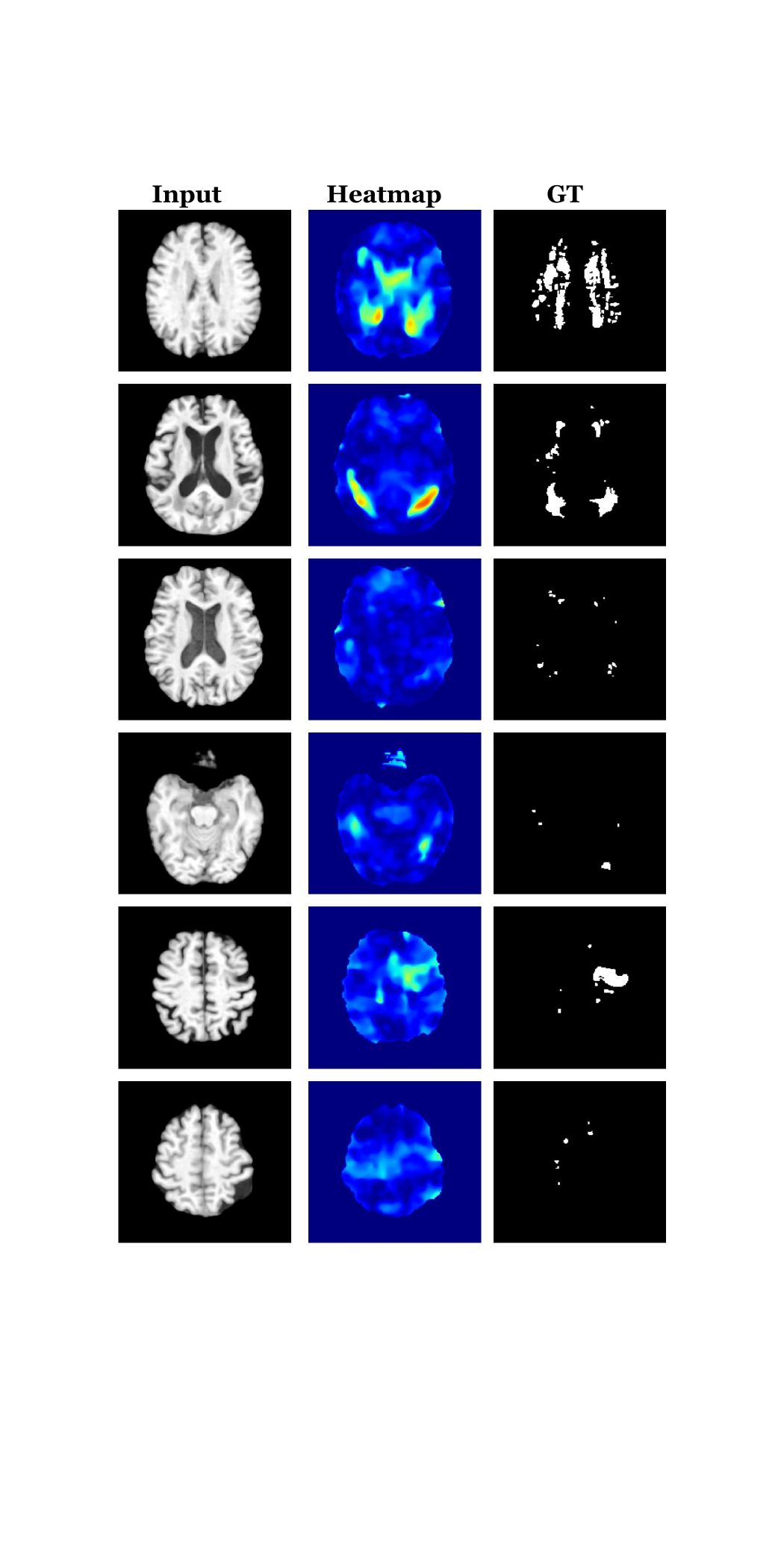}
\caption{Visualization of slices from WMH (T1). From left to right: the input slice, the heatmap (ranges from 0 (blue) to 12 (red)), and the ground truth for abnormal tissues.}
\label{fig:appendix_images_wmh}
\end{figure}

 Figure~\ref{fig:kde_all} shows kernel density estimates of the log-error and log-variance of normal and abnormal patches from the test set of IXI (both T1 and T2 modalities), ATLAS, MSLUB, and WMH datasets. 
Figure \ref{fig:appendix_images_ixi} shows slices from IXI. Figures  \ref{fig:appendix_images_brats},  \ref{fig:appendix_images_mslub}, \ref{fig:appendix_images_atlas}, and \ref{fig:appendix_images_wmh} show example neuroimages,  Patch2Loc's heatmap, and the ground truth for representative slices from BraTS (T2),  MSLUB (T2), ATLAS (T1), and WMH (T1) datasets, respectively.

\twocolumn%

\end{document}